# Datasheets for AI and medical datasets (DAIMS): a data validation and documentation framework before machine learning analysis in medical research

Ramtin Zargari Marandi[1], Anne Svane Frahm[1], Maja Milojevic[1]

[1] - Centre of Excellence for Health, Immunity and Infections (CHIP), Rigshospitalet, Copenhagen University Hospital, Denmark

## Highlights

DAIMS is a new framework for comprehensive data documentation and cleaning.

DAIMS focuses on medical tabular datasets ready for machine learning.

DAIMS allows selecting relevant machine learning analyses.

DAIMS allows automated data quality checks and validation.

The proposed framework promotes transparent reporting and reproducibility.

## Abstract

Despite progresses in data engineering, there are areas with limited consistencies across data validation and documentation procedures causing confusions and technical problems in research involving machine learning. There have been progresses by introducing frameworks like "Datasheets for Datasets", however there are areas for improvements to prepare datasets, ready for ML pipelines. Here, we extend the framework to "Datasheets for AI and medical datasets - DAIMS." Our publicly available solution, DAIMS, provides a checklist including data standardization requirements, a software tool to assist the process of the data preparation, an extended form for data documentation and pose research questions, a table as data dictionary, and a flowchart to suggest ML analyses to address the research questions. The checklist consists of 24 common data standardization requirements, where the tool checks and validate a subset of them. In addition, we provided a flowchart mapping research questions to suggested ML methods. DAIMS can serve as a reference for standardizing datasets and a roadmap for researchers aiming to apply effective ML techniques in their medical research endeavors. DAIMS is available on GitHub and as an online app to automate key aspects of dataset evaluation, facilitating efficient preparation of datasets for ML studies.

## Overview

Prior to data analysis, it is necessary to prepare datasets to meet certain quality and formatting standards and to provide data documentation as part of the data preparation process. A notable framework that has addressed data documentation is Datasheets for datasets by Gebru et al.[1], however it is not specific to medical datasets. In clinical research, there are established standards like those from the Clinical Data Interchange Standards Consortium (CDISC)[2] to enhance data interoperability and quality. Despite such efforts, there is a notable lack of dedicated research tools specifically designed for data documentation and screening prior to machine learning (ML) applications. When integrating diverse data sources and preparing datasets for ML, existing frameworks like CDISC often fall short in addressing their complexities[3]. This gap suggests an opportunity for developing a user-friendly data documentation and validation framework that would be straightforward to implement and use by researchers aiming to conduct ML analyses on medical datasets.

Data cleaning and validation is essential in the medical domain due to the presence of unstructured medical notes and inconsistencies in data imports. Physician scientists and other medical experts play a crucial role in extracting valuable information from these sources, transforming raw data into structured, harmonized, and clean datasets. This process provides documented datasets, informed by the basic information about the dataset and the cleaning processes undergone. However, due to the wide variety of data preparation requirements for ML analysis, several standardization steps may still remain unresolved.

To address this shortfall, we propose a new framework including an extensive data documentation form to clarify the available data and research questions to address, a software tool for data validation, as well as a flowchart as a roadmap to select ML analyses to address the research questions. Our solution is entitled the "Datasheets for AI and Medical Datasets (DAIMS)", a modified version of "datasheets for datasets" originally proposed by Gebru *et al.*[1]. The form for data documentation includes a questionnaire (see DAIMS_DatasetName_DDMMYYYY.docx on https://github.com/PERSIMUNE/DAIMS). It includes the original questions posited by Gebru *et al.* with our modifications and extensions that include replacing less familiar terminology with more commonly used terms in the medical field, enhancing definitions and instructions, and facilitating easier response to questions and documentation versioning. The form thus can serve as a datasheet for a single study and a live documentation for future studies where the dataset may change by including new variables or excluding some patient groups for example to address different research questions.

The form (DAIMS datasheet) includes detailed sections covering all aspects of the dataset lifecycle. It begins with clear "definitions and instructions", followed by a "motivation" section addressing the dataset's purpose, creators, and funding sources. The "composition" section provides details on data types, outcomes, subpopulations, missing data, and confidentiality. It also includes a "collection process" section, describing data acquisition methods, sampling strategies, and ethical reviews. The section of "preprocessing, cleaning, labeling" addresses several data standardization requirements including data duplicates and missing values. The "uses" section identifies past and potential applications while specifying unsuitable tasks and ethical considerations. The "distribution" section outlines data sharing mechanisms, licensing terms, and regulatory restrictions. Finally, the "maintenance" section includes instructions for versioning, updates, and long-term support, along with a space for "references" and "correspondence" to document contributors and sources.

In comparison to the original "datasheets for datasets", DAIMS introduces critical enhancements for requirements of medical research. It expands upon the original datasheet concept by addressing domain-specific needs, including strict data privacy measures, such as the handling of sensitive patient information and compliance with ethical standards like informed consent and General Data Protection Regulation (GDPR). The section for "preprocessing, cleaning, labeling" includes a checklist that is added as part of DAIMS contribution to itemize data quality requirements for ML analysis.

DAIMS datasheet also strengthens the data lifecycle by providing guidelines for dataset maintenance, including versioning, mechanisms for updates, and clear attributions for interdisciplinary collaboration. Its inclusion of a comprehensive glossary and detailed instructions enhances usability, making it accessible to medical experts and data scientists. The integration of domain-specific standards, such as international medical terminologies (e.g., SNOMED CT, ICD), further adds value by promoting interoperability and applicability in diverse research contexts. Collectively, these features improve the transparency, reliability, and ethical integrity of medical datasets for an efficient process of data delivery for ML applications and alignment with medical research standards.

The checklist as part of DAIMS provides a practical overview of each dataset's standardization requirements (see Table 1). Researchers can quickly assess and mark items that require further attention or are uncertain. The checklist is scored on a binary scale—items marked "done" receive a score of 1, while those labeled "not done" receive a score of 0. This scoring system allows for easy determination of the dataset's documentation completeness, with scores presented as #done/(#done+#not done), such as 5/24 indicating that five out of 24 items have been checked. The first 15 items can be checked by the DAIMS data validation tool, an open-source application that can be launched locally using Streamlit and Python, and the rest of the items must be checked manually. The tool is available to all researchers to download and use on their datasets on https://github.com/PERSIMUNE/DAIMS. Some items are left to be checked manually because of the technical complexities around them.

Table 1 The data cleaning checklist as a part of datasheets for AI and medical datasets (DAIMS).

| Mark if true | Mark if unsure | Item | Comment (if not done) |
|---|---|---|---|
| ☐ | ☐ | 1. Wide format: Each row is an instance, and each column is a variable (in addition to patient ID and target/outcome variable) | |
| ☐ | ☐ | 2. Each patient has a unique identifier (ID) | |
| ☐ | ☐ | 3. No Unicode character[1] | |
| ☐ | ☐ | 4. No duplicate rows and columns | |
| ☐ | ☐ | 5. First column is patient ID | |
| ☐ | ☐ | 6. Last column is the outcome variable or the main outcome in the case of multiple outcomes | |
| ☐ | ☐ | 7. No further separator (e.g., "," for numbers with four digits or longer such as 1,050,099) and no extra characters "()","[]", "<>", "//", "\|\|" and "{}". The only acceptable separator is the decimal point | |
| ☐ | ☐ | 8. All missing entries are indicated by the same entry (e.g., either "missing" or an empty field for not available) | |
| ☐ | ☐ | 9. No instance has all-missing variables | |
| ☐ | ☐ | 10. A data dictionary (codebook) is provided that defines all variables, their types (continuous, categorical or date and time) and units (e.g., kg) | |
| ☐ | ☐ | 11. The actual data values for continuous variables are within the range listed in data dictionary | |
| ☐ | ☐ | 12. Categorical variables have all their categories listed in the data dictionary | |
| ☐ | ☐ | 13. Rare categories in categorical variables are grouped | |
| ☐ | ☐ | 14. Perfectly collinear variables are removed | |
| ☐ | ☐ | 15. Irrelevant observations that could skew the results or cause bias (e.g., outliers or extreme values that do not reflect normal conditions) are removed | |
| ☐ | ☐ | 16. Date and time are formatted as described in the data dictionary (for example 05-11-2022 12:23 DD-MM-YYYY CET) | |
| ☐ | ☐ | 17. Numerical entries have "." as decimal separator (i.e., 1.2 not 1,2) | |
| ☐ | ☐ | 18. Non-English entries are translated to English. Non-Latin scripts are transformed to Latin scripts | |
| ☐ | ☐ | 19. The terms used in the dataset follow international standards[2,3,4] | |
| ☐ | ☐ | 20. All entries for each variable follow the same standard (e.g., older entries may have different standards or definitions) | |
| ☐ | ☐ | 21. Erroneous data are corrected or removed (e.g., a BMI value of 2000) | |
| ☐ | ☐ | 22. Informative missingness is properly encoded (e.g., "not tested" when it was deemed unnecessary to test) | |

| | | 23. Variables irrelevant for the study are removed (i.e., non-generalizable variables that could not relate to the outcome, e.g., billing ID or personal contact information) | |
|---|---|---|---|
| ☐ | ☐ | 24. No sensitive data including name, address, or identity number of the participants (patients) are included | |

[1] https://en.wikipedia.org/wiki/List_of_Unicode_characters
[2] https://en.wikipedia.org/wiki/MEDCIN
[3] https://en.wikipedia.org/wiki/International_Classification_of_Diseases
[4] https://en.wikipedia.org/wiki/SNOMED_CT

The checklist corresponds to another document, the "data dictionary", which provides an overview of variables that are present in the dataset (see Table 2). Having a comprehensive data dictionary is essential for clear communication and consistent usage of each variable within the dataset. By defining key elements such as "Variable Name", "Variable Units or Formatting", "Variable Type", "Variable Description", "Example", "Role", and "Typical Observation Error", the data dictionary facilitates seamless collaboration among team members from different backgrounds. "Typical Observation Error" is an important item that is usually not reported but is the same concept as measurement error and device accuracy as key information to calculate uncertainties[4,5], in predicted risks of a disease for each individual patient for example. All items except for the latter are necessary to conduct ML studies.

Table 2 An example of a data dictionary to characterize a dataset for machine learning analysis. Here you can find some examples of how different variables can be defined in the data dictionary.

| Variable Name | Units | Categories or range | Variable Type | Variable Description | Example | Role* | Typical observation error |
|---|---|---|---|---|---|---|---|
| Patient ID | Unitless | 000000; 999999 | Categorical** | Unique identifier for a patient | 123456 | Identifier | None |
| DOA | DD-MM-YYYY | 01-01-1900; 01-01-2025 | Date and time | Date of Admission | 01-01-2000 | Predictor | None |
| Death | YYYY | 1900; 2025 | Date and time | Date of death in year | 2020 | Outcome | None |
| Gender | | Male; Female | Categorical | Patient's gender | Male | Predictor | None |
| BMI | kg/m^2 | 15; 50 | Continuous | Patient's Body Mass Index | 25 | Predictor | ±0.2 kg/m^2 |
| Systolic blood Pressure | mmHg | 50; 250 | Continuous | Patient's systolic and diastolic blood pressure | 120 | Predictor | ±5 mmHg |
| Smoking Status | NA*** | Yes; No | Categorical | Whether the patient is a current smoker | Yes | Predictor | None |
| Diagnosis | Unitless | Negative; Positive | Categorical | Patient's diagnosis | Diabetes | Predictor | None |
| Treatment | Unitless | Treatment A; Treatment B | Categorical | Patient's treatment plan | Metformin 500mg daily | Predictor | None |
| Notes | Unitless | NA | Categorical | Additional notes about the patient | Patient has a history of hypertension | Other | None |
| HMG | g/dL | 11; 18 | Continuous | Hemoglobin level | 14 | Predictor | 2-5% |
| WBC | K/uL | 4; 11 | Continuous | White blood cell count | 7.5 | Predictor | 5-10% |
| Platelet Count | K/uL | 150; 450 | Continuous | Platelet count | 250 | Predictor | 5-10% |
| Creatinine | mg/dL | 0; 1.5 | Continuous | Creatinine level | 1.2 | Predictor | 5-10% |
| Glucose | mg/dL | 50; 200 | Continuous | Glucose level | 125 | Predictor | 5-10% |
| HbA1c | % | 0; 100 | Continuous | Hemoglobin A1c level | 7 | Predictor | 2-3% |
| Cholesterol | mg/dL | 0; 300 | Continuous | Total cholesterol level | 200 | Predictor | 5-10% |
| LDL | mg/dL | 0; 300 | Continuous | Low-density lipoprotein cholesterol level | 130 | Predictor | 5-10% |
| HDL | mg/dL | 0; 100 | Continuous | High-density lipoprotein cholesterol level | 60 | Predictor | 5-10% |
| Triglycerides | mg/Dl | 0; 600 | Continuous | Triglyceride level | 150 | Predictor | 5-10% |
| Microalbumin | mg/Dl | 0; 400 | Continuous | Microalbumin level | 30 | Predictor | 10-20% |

| CRP | mg/Dl | 0; 100 | Continuous | C-reactive protein level | 5 | Predictor | 5-10% |
| Staphylococcus aureus | NA | Positive; Negative | Categorical | Staphylococcus aureus infection | Positive | Predictor | None |
| Influenza A | NA | Positive; Negative | Categorical | Influenza A infection | Positive | Outcome | None |
| Disease stage | NA | Mild; moderate; severe; critical | Categorical | Disease stages | Mild | Outcome | None |
| HIV | NA | Positive; Negative | Categorical | Human Immunodeficiency Virus status | Negative | Outcome | Clinically confirmed |

* Depending on the domain, variables may have alternative descriptions. This table uses a single set of terms that are common for clarity and consistency; however, in various contexts, similar terms might be employed. ** Identifiers are not analyzed along with other categorical variables. *** NA: not available

The final component of DAIMS focuses on selecting appropriate ML methods to address specific research questions. To facilitate this process, we have developed a comprehensive flowchart diagram as a guiding framework, depicted in Figure 1. The design and ordering of the flowchart's blocks are grounded in common ML tasks prevalent within medical research. The flowchart starts with identifying the problem type, such as predicting numerical outcomes, categorical outcomes, or survival analysis. For binary or multiclass predictions, the chart distinguishes between tabular and image data, recommending interpretable models (e.g., logistic regression or decision trees) for tabular data and advanced architectures (e.g., ResNet[6], DenseNet[7], or U-Net[8]) for image data. If model performance needs to be improved, more complex models like tree-based ensemble models (e.g., LightGBM[9] and CatBoost[10]) are suggested. Text or time-series data, common in healthcare (e.g., Deoxyribonucleic acid - DNA sequences, electrocardiography - ECG signals), are directed towards specialized models like recurrent neural networks or graphical models[11]. If no labels are available, exploratory techniques such as clustering is suggested. For multimodal or complex datasets, model fusion techniques are advised[12]. These are some examples on how to navigate through the flowchart.

It is important to acknowledge that there is no universal solution for choosing ML models; instead, this flowchart serves as a foundational guide to navigate the diversity of available techniques. The emphasis on performance improvement within the flowchart is specifically directed towards optimizing tabular data processing, although its applicability extends to other data types such as images and text. Model evaluation and interpretation are also necessary steps that are not the focus of the flowchart.

Our approach in methodological choices is based on overall data and task characteristics as critical criteria to select ML tools. The flowchart is designed to be adaptable across various medical AI applications, promoting a standardized yet flexible framework for model selection and development. While this protocol provides a robust foundation, it acknowledges the evolving nature of ML techniques and encourages continuous evaluation and iteration based on emerging research and technological advancements. Future versions of the flowchart can be extended to incorporate emerging models and methods, like Large Language Models (LLMs)[13] and causal ML (e.g., RLearner[14]). Another important direction to extend the flowchart is to include methods for explainable AI such as SHapley Additive exPlanations – SHAP[15,16], model evaluation and validation, model safety and fairness approaches, as well as model deployment strategies. These are all important areas that can provide a comprehensive atlas for the implementation of ML models. The main limitation, however, would be the complexity of such a flowchart that covers all of these aspects. Therefore, the current version of the flowchart provides a balanced view of the most common approaches with a focus on clinical applications.

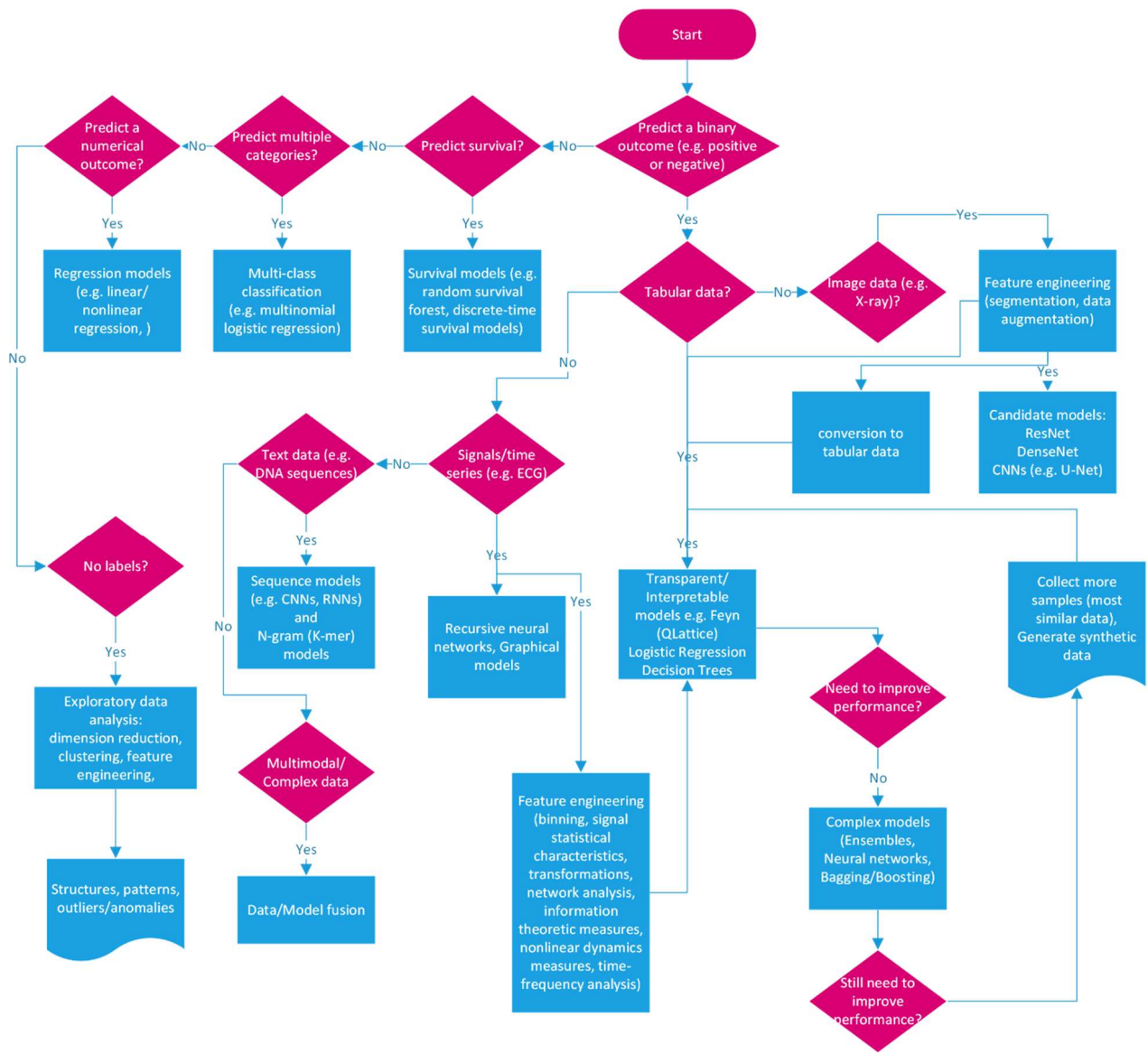

Figure 1 Flowchart to guide medical AI projects. The flowchart guides users through various pathways based on the type of data (e.g., numerical, categorical, image, text, time-series) and task (e.g., prediction, survival analysis, exploratory analysis). It highlights appropriate models and techniques, such as regression models, neural networks, and interpretable algorithms, while emphasizing iterative performance improvement through feature engineering, data augmentation, and ensemble methods. Although not explicitly mentioned, Large Language Models (LLMs) and NLP techniques are also highly relevant for text-based and multimodal data. While it provides an overview, detailed analysis steps like interpretation, evaluation, and deployment are not included for simplicity. The list of models and methods may not be exhaustive.

## Discussion

Overall, DAIMS not only enhance consistency in methodology within medical AI studies but also provide practical tools to organize data documentation and selection processes for ML methods. By leveraging DAIMS and its accompanying checklist, it is easier to navigate and apply appropriate ML analytical techniques, given that it covers a wide variety of research questions and solutions.

The original datasheet framework for datasets, proposed by Gebru *et al*.[1], offered a foundational structure for enhancing transparency and accountability in dataset documentation. However, it lacked domain-specific guidance critical for clinical and medical datasets, such as considerations for patient confidentiality. The current framework, DAIMS, addresses these gaps by incorporating domain-specific terminology, introducing a detailed checklist for data cleaning and standardization in medical data, and emphasizing subpopulation analysis and outcome variables. It also adds comprehensive sections on ethical review processes, data protection impact analyses, and mechanisms for dataset maintenance and updates, in line with medical research needs. These improvements make the framework more practical for documenting datasets in medical contexts, while still retaining its original focus on transparency and reproducibility.

Data integrity in clinical datasets is critical due to their complexity and regulatory requirements. In Python, tools like Pandas[17], NumPy[18], and Scipy[19] help identify missing values, duplicates, and outliers, while Matplotlib[20] and Seaborn[21] assist in visual anomaly detection. Frameworks like ydata-profiling[22] and Great Expectations[23] provide automated data profiling, with the latter requiring technical Python knowledge to define "expectations" (e.g., valid ranges, uniqueness, or type consistency) and detect data quality failures[23]. In R, the tidyverse suite of packages[24] supports data cleaning, transformation, and visualization. In contrast, DAIMS address practical challenges in data quality checks and documentation, tailored for medical research, while striving to decrease the reliance on complex technical processes.

While the flowchart does not explicitly mention LLMs, it is important to acknowledge their central role in many modern ML applications due to their versatility and adaptability. LLMs, such as GPT or transformer-based models like BioBERT[25], have become important in various domains, including healthcare in handling unstructured data like clinical text and patient histories. They bridge gaps between traditional feature engineering and deep learning by providing a novel framework for extracting relevant information and generating predictions for a wide variety of data types and modalities[26].

Although the current version of DAIMS addresses many critical issues that should be reviewed in a dataset prior to ML, it does not eliminate the need for additional checks. Researchers remain responsible for confirming that the dataset is properly formatted and suitable for the intended ML analysis. A crucial aspect of this preparation involves verifying that the dataset does not contain sensitive information, such as patient names, addresses, phone numbers, or other personal identifiers, to adhere to ethical guidelines. These checks are fundamental and should be conducted manually or using supplementary tools. Moreover, some ML pipelines, like medical artificial intelligence toolbox (MAIT)[27], include exploratory data analysis steps that assist in detecting and addressing several data-related issues. Therefore, even after following the current guide for data checks, it is highly recommended to use standardized and reliable ML pipelines to further identify and mitigate potential biases in the dataset.

In summary, we have proposed a new framework for data documentation and validation with structural improvements over previous solutions, notably the Datasheet for datasets", to conduct medical research studies where ML methods are applied. The improvements include extensions to the original datasheet with guidelines on how to fill it for medical studies, providing a checklist and a data dictionary together with a software solution to validate datasets, and a flowchart to find ML analyses for a wide variety of research questions. It also can serve as a documentation reference and framework for data versioning and warehousing where researchers can make informed decisions about the version of the dataset needed for a new study. DAIMS is also formulated to cover a broad set of research applications and that means it can be applied to any relevant research area and is not limited to ML or medical research only. The data validation and documentation framework can be utilized across diverse research and application scenarios. A key use case is reporting detailed dataset information as part of research publication, such as in Paetzold et al.[28]. DAIMS also supports open science initiatives by facilitating dataset sharing and can serve as a tool for grant proposals or regulatory submissions.

## Funding

The work was supported by the Danish National Research Foundation (DNRF126). The funders had no role in study design, data collection and analysis, decision to publish, or preparation of the manuscript.

## Ethical approval

There were no ethical concerns for this study.

## Data availability

Not applicable.

## Code availability

All code used in this study is available on the GitHub repository

https://github.com/PERSIMUNE/DAIMS.

Application is available online at:

https://daims-app.streamlit.app/

## Competing Interest

The authors declare no competing interests.

Neuroscience (VesselGraph). (2021) doi:10.5281/zenodo.5301621.